\documentclass[article,12pt]{article}
\usepackage{graphicx}
\usepackage{amssymb,amsmath,array,algorithm}
\usepackage{algpseudocode}
\usepackage{dcolumn}
\usepackage{bm}
\usepackage{dsfont}
\usepackage{color}
\usepackage{bbold}
\usepackage{geometry}
\usepackage{todonotes,braket,url}
\usepackage{mathrsfs}

\begin{document}

\title{Learning in Quantum Control: High-Dimensional Global Optimization for Noisy Quantum Dynamics}


\author{Pantita Palittapongarnpim$^1$,Peter Wittek$^{2,3}$,Ehsan Zahedinejad$^{1}$, \\
	Shakib Vedaie$^{1}$ and Barry C. Sanders$^{1,4,5,6}$}
%
%

\maketitle

\begin{center}
1- Institute for Quantum Science and Technology, University of Calgary \\
Calgary, Alberta T2N~1N4 Canada
%
2- ICFO-The Institute of Photonic Sciences \\
Castelldefels (Barcelona), 08860 Spain
\vspace{.1cm}\\
3- University of Bor\aa s \\
Bor\aa s, 501 90 Sweden
\vspace{.1cm}\\
4- Program in Quantum Information Science\\
Canadian Institute for Advanced Research\\
Toronto, Ontario M5G~1Z8 Canada
\vspace{.1cm}\\
5-Hefei National Laboratory for Physical Sciences at Microscale,\\ 
			University of Science and Technology of China,\\ 
			Hefei, Anhui 230026, People\textsc{\char13}s Republic of China
\vspace{.1cm}\\
6-Shanghai Branch, CAS Center for Excellence and Synergetic Innovation Center\\ 
			in Quantum Information and Quantum Physics,\\ 
			University of Science and Technology of China,\\ 
			Shanghai 201315, People\textsc{\char13}s Republic of China

\end{center}

		\begin{abstract}
Quantum control is valuable for various quantum technologies such as high-fidelity gates for universal quantum computing, adaptive quantum-enhanced metrology, and ultra-cold atom manipulation. Although supervised machine learning and reinforcement learning are widely used for optimizing control parameters in classical systems, quantum control for parameter optimization is mainly pursued via gradient-based greedy algorithms. Although the quantum fitness landscape is often compatible with greedy algorithms, sometimes greedy algorithms yield poor results, especially for large-dimensional quantum systems. We employ differential evolution algorithms to circumvent the stagnation problem of non-convex optimization. We improve quantum control fidelity for noisy system by averaging over the objective function. To reduce computational cost, we introduce heuristics for early termination of runs and for adaptive selection of search subspaces. Our implementation is massively parallel and vectorized to reduce run time even further. We demonstrate our methods with two examples, namely quantum phase estimation and quantum gate design, for which we achieve superior fidelity and scalability than obtained using greedy algorithms.
		\end{abstract}

	\section{Introduction}
	\label{sec:intro}
	Quantum mechanics has been recognized as a superior foundation for performing computation~\cite{Cav15,Shor97,Grov96}, secure communication~\cite{BB84,SBC+09} and metrology~\cite{GLM11,TA14}, also leading to technological advancements such as nuclear magnetic resonance and other resonators~\cite{NMR1,KRK+05,RSK05,MVT98,HJH+03}, femtosecond lasers~\cite{ABB+98,MS98} and laser-driven molecular reactions~\cite{BS92,TR85}. Central to these applications is the ability to steer quantum dynamics towards closely realizing specific quantum states or operations; i.e., the ability to control the system~\cite{DP10}.
	
	Control theory of classical systems has a long history and is extremely well developed. Control theory most often relies on a mathematical model of a physical system. The primary goal of control is to make the system's dynamics follow a reference trajectory or optimize the dynamics according to an objective function if a reference trajectory is not available. A mathematical model, however, can be difficult to specify exactly or solve analytically. Reinforcement learning is an alternative approach to control that does not necessarily have an explicit mathematical model of the underlying physical system, but rather it optimizes system's behavior by studying responses given a set of inputs~\cite{SBW92,KLM96}. If the control provided by the reinforcement learning algorithm is discrete in time, we can view the algorithm as a form of machine learning where the typical assumption of sampling from independent and identical distributions is replaced by that of a Markov decision process.
	
	The purpose of quantum control is identical to the classical case: generate a feasible {\it policy} for the given control problem. A policy is a set of instructions that determine the control parameters, and hence the effectiveness of the control scheme. This task is complicated by the quantum mechanical nature of the system, which allows non-classical correlations and noncontinuous jumps of the system's state~\cite{DP10}.
	For control tasks that involve continuous control over quantum states, usually through applying control pulses, algorithms such as GRadient-Ascent Pulse Engineering (GRAPE) have been applied to generate policies. These tasks are found in spectroscopy~\cite{KRK+05}, ultracold-atom research~\cite{KPK+04,JRG+14}, and implementation of quantum computation~\cite{RJ12}. When feedback is included, such as for adaptive parameter estimation~\cite{AJSD+02,WBB+09,Cap12} and for stabilization of a quantum state~\cite{MV07,VMS+12}, the dynamics of the state becomes nonlinear and noncontinuous. In this case, the optimization methods account for the quantum state’s allowed trajectories~\cite{WK98,BW00,WMW02,RPH2015}.
	
	As in the case of classical control, if the mathematical model of the quantum physical system is overly complex or elusive altogether, we can turn to reinforcement learning. For instance, in quantum control problems that involve measurement and feedback, reinforcement learning is gaining attention. Examples include an agent-based model in measurement-based quantum computation~\cite{TGB15}, mapping quantum gates on a spin system~\cite{BPB15}, suppressing errors in quantum memory~\cite{AN16}, optimization in ultra-cold-atom experiments~\cite{WEH+15}, and earlier work on the adaptive quantum phase estimation problem using heuristic optimization~\cite{HS10,LCPS13}. Whereas optimization methods such as Bayesian and Markovian feedback require the knowledge of the system dynamics~\cite{WMW02}, the machine learning approach~\cite{Bis06} enables us to treat the quantum system as a black box. The policy is generated in response to the outcome that closely approximates the target channel of the procedure irrespective of the dynamics involved. This approach has been used successfully to find policies for quantum control problems, such as the classification of qubits and trajectories~\cite{MGM+15,MW10}.
	
	Greedy algorithms are used for finding successful policies because the algorithms are fast in converging on successful solutions when performing local searches. Greedy algorithms are not guaranteed to succeed (i) when the search domain is non-convex or (ii) when the computational resource or the time for performing the control task is constrained~\cite{ZSS14}. Early quantum-control schemes employ standard heuristics, such as genetic algorithms, to find successful policies~\cite{SD94,BYW+97}. These algorithms fail when the number of control parameters increases or when decoherence and loss are included.
	
	As greedy algorithms fail to provide successful policies for the quantum control problems at hand, and simple heuristics for global optimizations also fail under realistic conditions of quantum systems, we develop new variants of optimization algorithms. We consider in particular differential evolution (DE) as a basis. Our decision is due to the algorithm's superior performance to particle swarm optimization (PSO) and other evolutionary algorithms for high-dimensional optimization problems~\cite{SP97,VT04,PSL05}.
	
	We demonstrate learning-based quantum-control schemes to find successful policies for two topics relevant to quantum control: phase estimation via adaptive quantum-enhanced metrology (AQEM) and designing fast quantum logic gates. Adaptive phase estimation aims to estimate an unknown phase shift on a light field such that the precision is enhanced by the use of a quantum state of light~\cite{BW00,RPH2015,Wis95,WK97,OIO+12}. Quantum metrology has many applications, such as in atomic clocks and gravitational wave detection. Quantum gate design tackles the problem of performing a specific gate operation on a quantum bit by steering its evolution~\cite{BPB15,ZGS15,ZGS16}. Fast quantum logic gates are required for designing fast quantum processing units as the timescales are limited by the decoherence time of the qubits. 
	Both problems require optimization of the procedure over the limited resource and time.
	
	Building on the DE-based reinforcement learning and machine learning algorithm, we address critical issues for applying quantum control in realistic physical systems in the following way.
	\begin{enumerate}
		\item We develop a scheme that can operate when practical imperfections such as noise and loss are included. The primary means to this is the way in which the objective function is evaluated.
		\item We improve scalability to a higher dimensional search space. For the problem of adaptive phase estimation, scalability is achieved by the accept-reject criterion that allows an early or a late termination of calculations. For the problem of gate design, we devise a subspace-selective self-adaptive DE (SuSSADE) that alternates between a search in the subspace and the overall space while adaptively updates the algorithmic constants of the standard DE algorithm during the search process.
		\item Furthermore, we vectorize the time-critical operations to use the parallel resources available efficiently in contemporary CPUs and GPUs.
	\end{enumerate}
	
	This article is structured as follows. In Sec.~\ref{sec:quantum control}, we introduce the relevant concepts in quantum mechanics and quantum control. We also explain the control procedures in our two examples: adaptive phase estimation and quantum-gate design. In Sec.~\ref{sec:machine learning}, we describe the connection between machine learning, reinforcement learning, and control. In Sec.~\ref{sec:method} we formulate adaptive phase estimation and quantum gate design as learning problems and show the methods for creating noise-resistant DE and increasing the scalability of our learning algorithms. The results for both control problems are in Sec.~\ref{sec:results}. 
	\section{Quantum control framework}
	\label{sec:quantum control}
	
	Quantum control concerns the application of control procedures to systems whose dynamics are governed by quantum mechanics~\cite{DP10}.
	In this regime, behaviors that are associated with classical dynamics are violated, leading to challenges in applying classical control theory directly to the system~\cite{DHJ+00, AT12}.
	In this section, we explain the key concepts in quantum mechanics that are necessary to understand quantum control, especially to adaptive phase estimation and quantum gate design.
	Readers interested in the complete formalism of quantum mechanics are referred to the many publications on this subject~\cite{Wat11, Hol12}.
	
	\subsection{Elements of Quantum Mechanics}
	
       In the regime of quantum mechanics, the state of an isolated particle A is given by a vector $\ket{\psi}_{\text{A}}$ of norm one in a Hilbert space $\mathscr{H}_\text{A}$.  
       We restrict our attention to the finite dimensional case where the Hilbert space is $\mathbb{C}^n$.    
    More generally, the state of a particle is a self-adjoint trace class operator of trace one~\cite{DFM-Z08}, given in the case of the adjoint vector $\ket{\psi}_{\text{A}}$ as $\hat{\rho}=\ket{\psi}_{\text{A}}\bra{\psi}$, called a density operator.
    The density operator can be represented by a matrix given for a chosen basis of the Hilbert space.
    
    Whereas a classical particle has definite values for its characteristics such as position and momentum, a characteristic of a quantum particle can only be described in terms of its probability distribution.
    Given a chosen basis $\{\ket{i}_\text{A}\}$, the state can be represented by $\ket{\psi}_{\text{A}}=\sum\limits_{i} c_i\ket{i}_\text{A}$, where $c_i\in\mathbb{C}$.
    The absolute square of $c_i$ determines the probability distribution on the chosen basis and, hence, must satisfy the condition $\sum\limits_{i}\left|c_i\right|^2=1$.
    In the matrix representation of the density matrix, the distribution is determined by the diagonal elements, which leads to the trace of one.      
    
	When two or more particles exist in the system, correlations can exist that cannot be expressed by any local hidden variable model.
	For instance, a qubit is a state in $\mathbb{C}^2$, and a two-qubit state of particles A and B is in $\mathbb{C}^2_\text{A}\otimes\mathbb{C}^2_\text{B}$.
	A quantum correlation between the qubits leads to the state $\ket{\psi}_\text{AB}$ that cannot be expressed as $ \ket{\psi}_{\text{A}}\otimes\ket{\psi}_\text{B}$ and a phenomenon known as entanglement where a local operation on the subsystem $A$ affects the state of the subsystem $B$ regardless of the space separation between $A$ and $B$.
    
    The connection between state and probability distribution becomes apparent when we consider the measurement of the state. Measurements are described by a positive operator-valued measure (POVM) $\{\hat{E}_x\}$~\cite{Bra03} acting on system A, which delivers outcome $x$ with probability $\operatorname{tr}\left(\hat{E}_x\ket{\psi}_\text{A}\bra{\psi}\right)$. POVMs satisfies the condition $\sum\limits_x \hat{E}_x^\dagger \hat{E}_x=\mathds{1}$, and $\hat{E}_x$ is positive for all $x$. 
    The state after the measurement is
    \begin{equation}
     \ket{\psi,x}_\text{A}=\frac{\hat{E}_x\ket{\psi}_\text{A}}{\sqrt{\left|\hat{E}_x\ket{\psi}_\text{A}\right|^2}}.
     \end{equation} As the outcome $x$ is random, the state after the measurement is a random jump unless the state is an eigenstate of $\hat{E}_x$.

    
    Manipulation of the state is accomplished by providing outside interaction to the system. This interaction is described by a completely positive trace-preserving (CPTP) map~\cite{Hol12}, ensuring that the vector remains a quantum state. We refer to these maps as quantum channels $\{\hat{\mathcal{C}}_j\}$ that satisfy $\sum\limits_{j}\hat{\mathcal{C}}_j\hat{\mathcal{C}}^\dagger_j=\mathds{1}$. If these linear operators are unitary, the system does not interact with its environment, and its dynamics remains reversible.
    
    As particles, in reality, are not perfectly isolated, the system can be considered as having a constant weak interaction with a bath, which is another quantum system in a larger Hilbert space.
    The state of the bath is not accessible through measurement, and this loss of information leads to the decoherence of the system's state.
    Quantum correlations suffer from this kind of interaction because the state turns into a convex combination of the resulting states of all possible interactions between bath and system, $\hat{\rho}_\text{A}=\sum\limits_{i}P_i\ket{\psi^{(i)}}_\text{A}\bra{\psi^{(i)}}$, where $\{P_i\}$ is a classical probability distribution. The state is no longer pure, but forms a mixed state. In the case of strong system-bath interaction, the system loses the quantum-mechanical characteristics altogether~\cite{Zur07}.
    
    The essence of quantum control is to steer a quantum channel towards the desired operator.
    One way to test the channel resulting from the control is to monitor the channel through process tomography~\cite{DPS03}. A known quantum state is injected as the input, and the output state is measured. This random measurement outcome is then used to infer the operation performed by the channel. In the process of obtaining a successful policy, the outcome is monitored, and the policy that is used to adjust the control parameters is updated accordingly. The difference between the channel and the target is determined using an objective function. However, if the goal of the control does not explicitly involve the channel, other methods and objective functions can be selected.
    
    In this work, we consider two examples of quantum-control procedures. The first example is the adaptive quantum-enhanced metrology, which utilizes a multiparticle entangled state to attain quantum-enhanced estimation of an unknown parameter and consolidates previous material~\cite{PWS16}.
    The second case study is quantum gate design, which uses the control to apply logic gates on three quantum bits (qubits)~\cite{ZSS14}.
	
	\subsection{Adaptive quantum-enhanced metrology}
	\label{AQEM}
    The task of a quantum-enhanced metrology (QEM) scheme is to infer an unknown parameter $\phi$ using entangled states of $N$ particles such that the scaling in uncertainty surpasses $\Delta\phi\propto1/\sqrt{N}$ obtained using a classical strategy~\cite{TA14}. This scaling is known as the standard quantum limit (SQL). The use of quantum resources enables a QEM scheme to approach the Heisenberg limit (HL) corresponding to the scaling of $\Delta\phi\propto1/N$~\cite{ZPK12}. This quadratic improvement in precision is valuable for applications where the measurement is operating at the limit of $N$ that can be safely produced or detected.
    
    AQEM is one strategy for performing QEM that involves splitting the input state into a sequence of single-particle bundles~\cite{BWB01}. A bundle is injected into the channel and measured at the output. The measurement outcome is used to update the control parameter in preparation for the next bundle. The value of the control parameter after the $N^\text{th}$ measurement is then taken to be the estimate.
    
    The evolution of the quantum state over the course of the measurement process is noncontinuous, and so the policy that can achieve quantum-enhanced precision is non-trivial to find. For this reason, machine learning has been introduced~\cite{BW00,HS10,LCPS13}. In our work, we focus on the channel that includes noise and loss, which are imperfections present in every practical measurement setup.
	
	\begin{figure}
		\centering
		\label{fig:MZIControl}
		\includegraphics[scale=0.35]{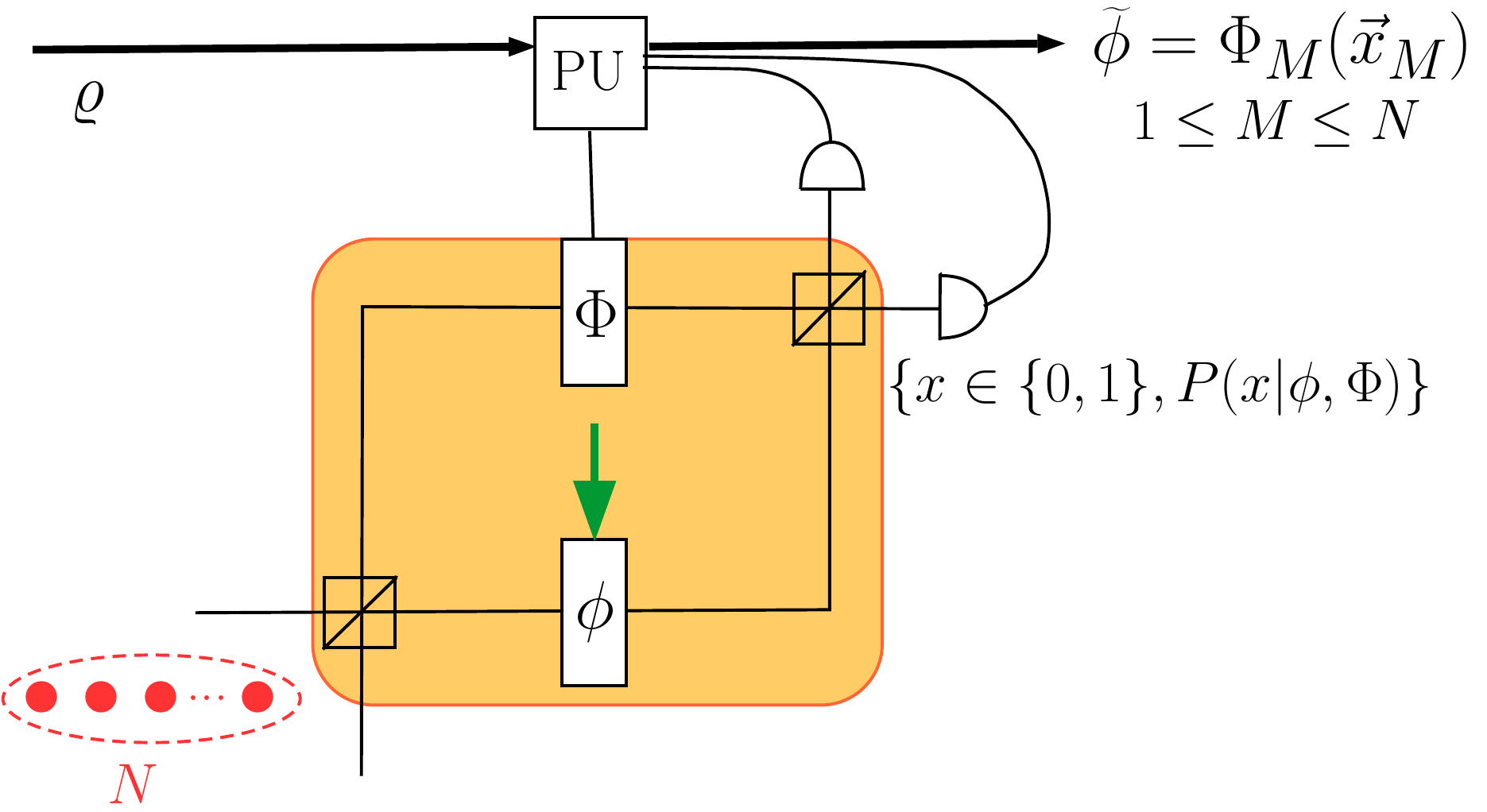}
		\caption{An adaptive phase estimation scheme. 
		An $N$-photon entangled state is separated into single-particle bundles. 
		A particle is injected into a Mach-Zehnder interferometer, representing a two-mode interferometer. 
		Contained in this interferometer is an unknown phase shifter $\phi$ and a controllable phase shifter $\Phi$, whose value after the $m^{\text{th}}$ measurement is $\Phi_m$. 
		The path $x_m$ in which the photon exits is detected by single-photon detectors connected to a processing unit (PU). 
		The PU uses the information to adjust $\Phi_{m-1}\to\Phi_{m}$ according to policy $\varrho$. After all the photons are measured, the estimate $\widetilde{\phi}$ is inferred from $\Phi_M$, allowing for loss in photons which leads to $1\leq M\leq N$.
		Hence, the estimate is a function of the history of measurement outcomes $\bm{x}_M=(x_1x_2\cdots x_M)$.}
		
	\end{figure}
	We consider the problem of optical interferometric-phase estimation (Fig.~\ref{fig:MZIControl}),
	which is well-studied due to its connection to the detection of gravitational wave \cite{Cav81,AAA+16} and atomic clocks~\cite{BS13}.
	The interferometer has two input modes and two output modes.
	The input state containing $N$ entangled photons is injected into the interferometer one photon at a time.
	
	Neglecting loss, the $m^\text{th}$ photon comes out from either of the output modes with a probability that depends on $\phi-\Phi_{m-1}$.
	Our interferometer model includes Gaussian noise on the phase shift with standard deviation $\sigma$.
	We label the outcome by $x_m\in\{0,1\}$, where 0 refers to the photon exiting the first port and 1 to the photon exiting the second port.
	The sequence of outcomes from the first to the $m^{\text{th}}$ photon is given by $\bm{x}_m=(x_1x_2\cdots x_m)$.
	
	The exit path of the photon is used to determine the value of $\Phi_m$ for the next round of measurement.
	Once all photons are put to use in the $M^{\text{th}}$ measurement, allowing for the loss of photons such that $1\leq M\leq N$, the estimate $\widetilde{\phi}$ of $\phi$ is inferred from $\Phi_M$ to be $\widetilde{\phi}\equiv\Phi_M$.
	As the measurement outcome $\bm{x}_{M}$ is a string of discrete random variables, the estimate of phase from this scheme, which is a function of $\bm{x}_M$, is also discrete.
	
	Because the distribution of the estimate $\widetilde{\phi}$ is periodic, the standard deviation is not an appropriate choice to quantify the imprecision $\Delta\widetilde{\phi}$. 
	Unless the domain is bounded, the standard deviation is skewed by the existing peak appearing in the distribution outside of the domain $\left[0,2\pi\right)$~\cite{Hil02}.
	The imprecision is instead quantified by the Holevo variance~\cite{BW00},
	\begin{align}
		\label{eq:Holevo variance}
		V_H & = S^{-2}-1,\\
		\label{eq:Sharpness}
		S & = \left|\sum\limits_{\widetilde{\phi}} P(\widetilde{\phi}|\phi) \text{e}^{\text{i}(\phi-\widetilde{\phi})}\right|,
	\end{align}
	which is one possible choice for a periodic distribution~\cite{FA01}. 
	Our goal is to generate a feedback policy such that $V_H$ is minimized and the power-law scaling with $N$ exceeds $1/\sqrt{N}$.
	
	\subsection{Quantum gate design}
	\label{QGD}
	
 Quantum computing employs quantum mechanics to perform computation and promises speed-up in computational time for algorithms such as factorization~\cite{Shor97} and database search~\cite{Grov96}. 
Implementing quantum computing has been challenging due to the interaction between the quantum system and the environment, which introduces errors and may even nullify the advantage of using quantum resources altogether~\cite{Shor96}. 
    If the error rate can be reduced to a value that is less than a specified threshold, error correction can be introduced, and quantum information is thereby protected~\cite{SWS16}.
    
Information is encoded on qubits, for each qubit has the state spanned by the basis $\left\{\ket{0},\ket{1}\right\}$. Therefore, unlike a classical bit, a qubit can exist in any superposition $a\ket{0}+b\ket{1}$, where $a,b\in\mathbb{C}$ and $\left|a\right|^2+\left|b\right|^2=1$. Quantum algorithms use these qubits as resources to perform computations. Just as in classical computation, the quantum algorithm operating on a large number of qubits can be decomposed into gates acting on a few qubits at a time. An ideal quantum gate is reversible and is represented by a unitary transformation $U$. For each of these gates, error threshold can be assigned such that fault-tolerant quantum computing is attained.
    
\begin{figure}
		\centering
		\includegraphics[scale=0.5]{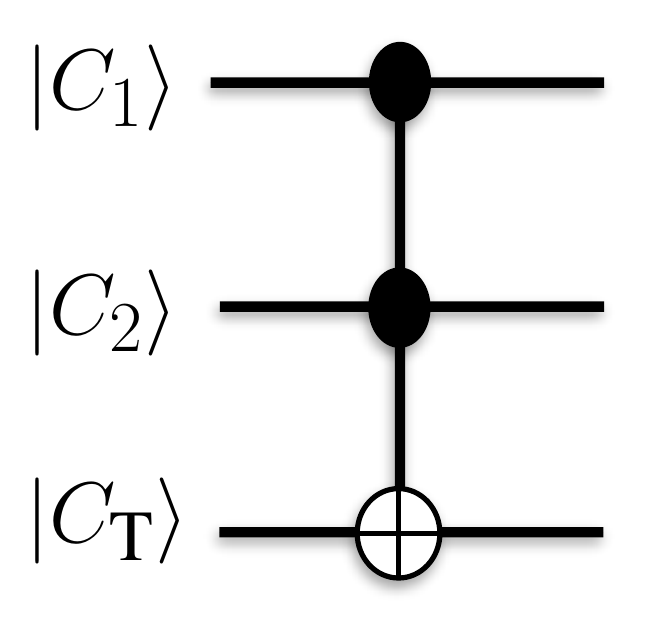}
		\caption{The quantum circuit representation of the Toffoli (CCNOT) gate. The horizontal solid black lines are circuit wires with each wire representing a qubit. {\large${\bullet}$} shows the control qubits and
			$\bigoplus$ denotes the NOT operator acting on the target qubit. $\ket{C_1},\ket{C_1},\ket{C_\text{T}}$ refer to the states of the first, second and target qubits.
			The gate accepts the input from the left side and output a new state on the right side.}
		\label{fig:Toffoli_CZ}
	\end{figure}

Quantum computing operations can be decomposed efficiently down to a set of one- and two-qubit gates~\cite{BBC+95,MVBS04}. 
However, this decomposition results in an increased processing time and lower overall fidelity. One way to increase the efficiency is to convert these operations to gates that act on more than two qubits. The Toffoli gate is one such gate. This gate is a controlled-controlled-not gate acting on three qubits (Fig.~\ref{fig:Toffoli_CZ}), whose action is summarized in Table~\ref{table:CCNOTtruth}.
	
	\begin{table}[t]
		\centering
		\begin{tabular}{c|c|c|c|c|c}
			\multicolumn{3}{c|}{Input} &
			\multicolumn{3}{c}{Output} \\ \hline
			$C_1$ &$C_2$& $C_T$ & $C_1$& $C_2$& $C_T$    \\\hline
			$\ket{0}$  &$\ket{0}$ &$\ket{0}$ &$\ket{0}$&$\ket{0}$&$\ket{0}$  \\
			$\ket{0}$  &$\ket{0}$ &$\ket{1}$ &$\ket{0}$&$\ket{0}$&$\ket{1}$  \\
			$\ket{0}$  &$\ket{1}$ &$\ket{0}$ &$\ket{0}$&$\ket{1}$&$\ket{0}$  \\
			$\ket{0}$  &$\ket{1}$ &$\ket{1}$ &$\ket{0}$&$\ket{1}$&$\ket{1}$  \\
			$\ket{1}$  &$\ket{0}$ &$\ket{0}$ &$\ket{1}$&$\ket{0}$&$\ket{0}$  \\
			$\ket{1}$  &$\ket{0}$ &$\ket{1}$ &$\ket{1}$&$\ket{0}$&$\ket{1}$  \\
			$\ket{1}$  &$\ket{1}$ &$\ket{0}$ &$\ket{1}$&$\ket{1}$&$\ket{1}$  \\
			$\ket{1}$  &$\ket{1}$ &$\ket{1}$ &$\ket{1}$&$\ket{1}$&$\ket{0}$  \\
		\end{tabular}
		\caption{The truth table representation of Toffoli gate. $C_1$ and $C_2$ denote the control qubits, and $C_\text{T}$ represents the target qubit. The Input and Output columns show the states of the three qubits before and after applying Toffoli.}
		\label{table:CCNOTtruth}
	\end{table}
	
	The current experimental schemes to design fast Toffoli gate are limited to decomposition approaches, with the fidelity limited to 68.5\% in a three-qubit circuit QED system~\cite{FSB+12}, 71\% in an ion-trap system~\cite{MKH+09}, 78\% in a four-qubit circuit QED system ~\cite{RDN+12} and 81\% in a post-selected photonic circuit~\cite{LBA+09}. Here our goal is to devise a machine learning algorithm to design a single-shot threshold-fidelity Toffoli gate without any need to resort to decomposition approach. Our approach to creating a high-fidelity Toffoli gate is not restricted to a specific physical model for quantum computation. We choose to design a Toffoli gate for an architecture of three nearest-neighbour capacitively coupled superconducting transmons~\cite{SHK+08}.

In a transmon system, the computational basis $\left\{\ket{0},\ket{1}\right\}$ is assigned to the two lowest energy levels, although the transmon system consists of multiple discrete levels. The evolution of the quantum states for the three transmons are controlled through electrical pulses changing the frequency of each of the transmons. Because the three transmons are coupled, these changes tune their interactions and allow us to perform gate evolution on the computational bases of the three qubits.
	
The unitary transformation of the Toffoli gate is approximated using a sequence of constant control pulses lasting $\delta t=1~\text{ns}$ applied over time $\tau$. At time $t$, the control pulses $\bm{\varepsilon}(t)=\left(\varepsilon_1(t),\varepsilon_2(t),\varepsilon_3(t)\right)$ is applied to the three transmons, effectively subjecting the system to a unitary transformation $U(\bm{\varepsilon}(t),\delta t)$. The approximate unitary for the gate is, therefore,
	\begin{equation}
		\tilde{U}(\bm{\varepsilon}_\tau,\tau)=U(\bm{\varepsilon}(\tau-\delta t),\delta t)U(\bm{\varepsilon}(\tau-2\delta t),\delta t))\cdots U(\bm{\varepsilon}(0),\delta t),
	\end{equation}
	where $\bm{\varepsilon}_\tau$ denotes the entire sequence of the three control pulses.
	
The gate's performance is quantified using the intrinsic fidelity~\cite{ZGS16},
	\begin{equation}
	\mathcal{F}=\frac{1}{8}\left|\text{tr}\left(U_{T}^{\dagger}\tilde{U}(\bf\varepsilon(\tau),\tau)\right)\right|,
	\label{eq:fidelity}
	\end{equation}
	where $U_T$ is the unitary transformation of the Toffoli gate. The fidelity $\mathcal{F}$ has a value between 0 and 1, where $\mathcal{F}=1$ is attained when $\tilde{U}=U_T$. We aim for an intrinsic fidelity beyond 0.9999 for the design of the Toffoli gate as required by fault-tolerant quantum computing.
	
In the experimental realization, non ideal behavior arises due to electronic imperfections.
        One is the distortion of the control pulses, which we treat as sequences of piecewise constant functions.
        A more realistic pulse shape takes into account the response time of the electronics, which acts as a Gaussian filter. 
        Another source of imperfection is the disturbance of the pulses associated with the thermal noise in the electronics.
        Although this noise is not included in the optimization, we test the robustness of our control pulses by adding random noise $\delta\varepsilon \cdot \text{rand}(-1,1)$, where $\text{rand}(-1,1)$ uniformly generates a random number in $(-1,1)$ to the control parameters at each time bin. The value of $\delta\varepsilon$ is varied from 0 to 300 kHz. We then use the distorted pulse to calculate the intrinsic fidelity for each value of $\delta \varepsilon$.
			
	Note that we have already devised a quantum control scheme to design a high-fidelity quantum gate~\cite{ZGS15,ZGS16}. Our main goal in this article is to present the problem in a different framework, namely machine learning. In particular, we formulate the problem for supervised learning. Expressing the problem in the machine learning context can provide a new perspective on finding control pulses for when the transformations, $U(\varepsilon(t),\delta t)$, are no longer unitary.
	
	\section{Machine Learning and Evolutionary Algorithms for Control}\label{sec:machine learning}
	
In this section, we explain how machine learning and reinforcement learning can be used as tools for control. 
A control problem involves optimization of a control policy such that a target performance is met~\cite{Leigh04}.
Optimizing based on a model is not useful if the model is incorrect, for example by not properly incorporating environmental interaction.
    As a satisfactory noise model might not exist, the alternative is to implement optimization procedures that are independent of the underlying interaction, with evolutionary algorithms being one such example.
    
    Control theory concerns finding a way to steer a system to achieve a target~\cite{Leigh04}. The method of control may involve monitoring of a system, in which case the control may involve a feedback loop adjusting the control signal following a policy. Such a control procedure is known as closed-loop control~\cite{Haid13}. For a system that cannot be monitored, or for a system whose monitoring is deemed unnecessary, an open-loop strategy is applied, and the control signal is predetermined using a model of the evolution of the system~\cite{Leigh04,AM08}.

For the cases where the system models are not known, not accurate, or too complicated to be used for generating feasible control policies, machine learning and reinforcement learning can be used to generate the policy through trial and error.
    Machine learning typically assumes sampling from random variables that are independent and identically distributed (i.i.d.). Concentration theorems based on this assumption give guarantees of estimation accuracy or generalization performance~\cite{Vap95,DGL96}. A variant of machine learning takes training instances one by one and updates its model accordingly. This scheme is known as online learning. Some results from statistical learning theory extend to this case if the samples remain i.i.d.~\cite{Long99}.
    
In reinforcement learning, an agent responds to a random outcome from its interaction with an environment~\cite{SB98}. The agent is equipped with a set of possible actions and a scoring function with which it can evaluate its performance. The agent can either greedily optimize the scoring function or aim to maximize long-term performance quantified after the task is completed.
    In this case, due to the interaction between the learning agent and the environment, the sampling loses the i.i.d. assumption. The most we can assume is that the interaction is modeled by a Markov decision process. 
    
    In both machine learning and reinforcement learning, we often boil down the learning problem to a constrained optimization problem with an objective function that can be regularized. The function is seldom convex, making it hard to find the global optimum of the problem. We can use a relaxation and a convex approximation or substitution. For instance, support vector machines replace the optimal 0-1 loss with the hinge-loss to have a convex optimization problem to solve~\cite{EBG11}. Another option is to use a non-convex optimization algorithm. This approach works if a convex relaxation is not easy to derive or when we are not satisfied with a local optimum. Sometimes even a greedy algorithm works well for the non-convex case, depending on the topology of the space.
    
    Derivative-free heuristics are common in global optimization. Two widely used algorithms are PSO~\cite{KE95}, which is a form of swarm intelligence, and DE~\cite{SP97}, which is an evolutionary algorithm. In comparative studies, DE has been shown to be the most powerful in that it converges on a near-optimal solution quickly and can find solutions without stagnation in higher dimensional search spaces than other algorithms~\cite{VT04}.
    
Quantum control tasks also employ closed-loop and open-loop controls to accomplish the desired goals.
    A key difference between quantum and classical control lies in the quantum system's response to the measurement.
    Whereas classical systems are unaffected by the measurement procedure, the quantum states response to a measurement with a random jump depending on the measurement outcome.
    For applications where the quantum states are important to the goal, such as gate design, measurements are to be avoided, and the control procedure for the task is designed based on open-loop control.
    On the other hand, when a quantum state is used as a resource for accomplishing the desired goal, or if measurements are possible, closed-loop control can be applied.

  To implement quantum control in the experimental settings, the control procedures must be designed to be resilient to imperfections, including the nonideal evolution of the quantum state due to its interaction with the environment. 
    Although noise models can be incorporated into computer simulations of the control procedures, the models might not match the noises in the real-world system.
    One approach is to design a procedure that can learn the control policies directly from experiments, including as few assumptions about the noise as possible.

    Reinforcement learning is a valuable tool for applying control as it provides a method to generate a policy that does not rely on the knowledge of the system~\cite{SBW92,KLM96,CDL+2014}. Learning is performed through trial and error, and the system's dynamics is treated as a black box. In fact, this black box treatment of a process is not restricted to a control scenario: if we only want to approximate the output of the process and we have a certain number of disposal of the black box, we can employ a supervised learning algorithm with the assumption of the sampling being i.i.d. For quantum control, this feature is valuable because the knowledge of the dynamics of a real quantum system is never exact. Furthermore, in systems involving measurements, the back-action leads to an exponentially-growing number of state trajectories with the number of possible measurement outcomes $\left|\{x\} \right|$, making the problem difficult to solve analytically.
	
	\section{Method}
	\label{sec:method}
	We now explain how to construct quantum-control procedures for AQEM and quantum gate design as learning problems and the challenges they pose to learning algorithms. We then explain how we create noise-resistant differential evolution and methods for generating policies in high-dimensional problems. Here we assign a fixed mutation rule, although a new DE variant created by random selection of the mutation is possible~\cite{MCD2015}.
	
	\subsection{Quantum control procedures as learning problems}
	
	In this subsection, we discuss how the control procedures are performed in an adaptive phase estimation scheme and on the superconducting circuit to create a Toffoli gate. The feedback control applied in the adaptive measurement fits into the framework of reinforcement learning, for which we discuss the policy and the training set. 
	Quantum gate design, on the other hand, fits with open-loop control as direct monitoring of quantum states interfere with their evolution.
	To move beyond treating the problem as optimization, in which the noise models are required, we formulate the gate design as a learning problem fit for supervised learning.
	
	\subsubsection{Generating feedback policy for AQEM}
	
    The goal of adaptive phase estimation is to infer the value of an unknown parameter $\phi$ following a sequence of measurements performed on a string of $N$ single-photon pulses. 
    In order to deliver quantum-enhanced precision, not only must the particles be entangled, but a policy must be generated that can attain the desired precision.
    In principle, the policy must be optimized for all sequences $\bm{x}_M$ and all value of $\phi\in\left[0,2\pi\right)$, which is impossible as $\phi$ is in a continuous domain.
    For this reason, we employ the reinforcement learning approach to generate a feasible policy using a training set selected from the domain of $\phi$.
    
    Reinforcement learning is the suitable framework because the approach is designed to optimize a decision-making process given random inputs.
    The adaptive phase measurement involves decision making by the processing unit in response to the random measurement outcomes, which fits in this framework.
    In the learning algorithm, the adaptive procedure is simulated many times with a fixed input quantum state and treated by the learning algorithm as a black box.
    The fitness values of the policies are computed over a training set of randomly generated $\phi$ and are used by the algorithm for the optimization.
        
    In AQEM, the feedback policy is a set of rules that determines how the controllable phase shifter $\Phi$ is adjusted.
    In the $m^{\text{th}}$ round of measurement, the policy is a function of the sequence of previous outcomes $\bm{x}_{m-1}\in\{0,1\}^{m-1}$.
    The process is better understood by representing a policy as a binary decision tree, where each branch from the root to the leaf corresponds to a sequence of $\bm{x}_N$.
    An advantage of using this representation is that the size of a policy is readily determined from the number of branches, and its size scales as $2^N-1$.
    
    The exponential scaling of the policy size makes generating the policy increasingly expensive in computational time,
    which limits the number of particles $N$ in which the application of reinforcement learning is practical. 
    To reduce the policy's size and make searching for the policy tractable, we restrict to Markov feedback~\cite{WO12}, 
    in which only the current outcome $x_m$ is used to determined the value of $\Phi_m$.
    In particular, we impose the update rule
    \begin{equation}
        \Phi_m=\Phi_{m-1}+(-1)^{x_m}\Delta_m;
        \label{eq:update}
    \end{equation}
    thus the decision tree is parameterized by a vector $\bm{\Delta}=\left(\Delta_1,\dots,\Delta_N\right)$.
    Hence, the size of the policy $\varrho=\bm{\Delta}$ is reduced to scale linearly with $N$.
    The space in which the policy can be searched is therefore restricted to $[0,2\pi)^N$.
    Through the implementation of this rule in previous work~\cite{HS11b,HS10,LCPS13}, we found that the rule leads to feasible policies for adaptive phase estimation.
    
    The update rule (\ref{eq:update}) restricts the estimate $\widetilde{\phi}\equiv\Phi_M(\bm{x}_M)$ to a discrete value even though $\phi$ is continuous in $\left[0,2\pi\right)$.
    In other words, the AQEM scheme delivers a discrete approximation of $\phi$ based on the discrete outcomes $\bm{x}_M$.
    As $N$ increases, the approximation becomes refined, and an increasingly precise estimate may be obtained.
    
    To determine the performance of a policy, the imprecision of all possible value of $\phi\in\left[0,2\pi\right)$ has to be taken into account. 
    This task poses an additional challenge to calculating Eq.~(\ref{eq:Sharpness}), which is computationally expensive.
    We instead estimate the value of $S$ from
    \begin{equation}
        S:=\left|\sum\limits_{k=1}^{K}\frac{\text{e}^{\text{i}\theta_k}}{K}\right|
        \label{eq:Sharpness_est}
    \end{equation}
    where $\theta_k=\phi_k-\widetilde{\phi}_k$ and $K=10N^2$.
    This particular choice of $K$ has been shown to deliver the estimate of $S$ that converges in the previous work~\cite{HS10}.
    The samples are also training data $\{\phi_k\}$ chosen uniformly from $[0,2\pi)$ to avoid the problem of overfitting.
    
    The adaptive phase estimation scheme works by refining the estimate $\widetilde{\phi}$ through tweaking $\Phi$ according to the outcome.
    In the event of small probability $\eta$ that the photon is lost, the automated system is instructed to do nothing in the absence of information to update the estimate.   
    We assume a small loss in which case we optimize without loss but test the performance accounting for loss.
    If the test succeeds, we adopt the policy; if the test fails, we repeat the optimization process.
	
	\subsubsection{Shaping the frequencies of transmon system}
	
	The goal for quantum gate design is to generate a set of pulse sequences $\bm{\varepsilon}_\tau$ such that the transformation of the three transmons over time $\tau$ is approximately the Toffoli gate.
	The pulses are affected by Gaussian distortion, thereby smoothing out the constant piecewise functions.
	Disturbances are included by adding frequency noises to the pulses after the optimization to test for robustness. 
    For both noises considered, the assumption of unitary transformation holds, and the problem of finding the control pulses can be treated as an optimization problem.
    In this work, however, we formulate this problem for supervised learning as the learning approach can be implemented regardless of whether the transformation is unitary or not.
      
    
   
        When the transformations of the quantum state are unitary, i.e., the quantum system is perfectly isolated from the environment,
        the control pulses can be optimized over a set of basis states.
        That is because the quantum state remains a pure state, and due to linearity in quantum mechanics, any transformation on the superposition of the basis states are equal to the superposition of the transformed basis.
        Hence, optimizing the gate's performance over the basis states is equivalent to the optimization of the problem over the entire state space.
        
        This observation is no longer true when the state becomes mixed, as is the case when the quantum system is coupled to the environment, and the transformations are represented by quantum channels.
        In this case, optimization over the basis states would not guarantee that the gate's performance is maximized over the entire space of all input states. 
        Hence, the optimization process has to take into account other input states as well. 
        Finding a feasible policy, in this case, is non-trivial as there are infinitely many superposition states.
        Machine learning becomes a promising approach for generating the control pulses. 
                
        The task of supervised learning is to discover a model given a set of input-output data (a training set), which approximates the true function that maps the inputs data to their corresponding outputs.
        By casting the Toffoli gate as a model for we wish to learn the parameters $\bm{\varepsilon}_{\tau}$, we formulate the Toffoli gate design as a regression problem.
        We use the basis states and their corresponding output states in Table~\ref{table:CCNOTtruth} as the training set.
        The time $\tau$ for the gate operation is found through trial and error such that the confidence in the model, quantified by the intrinsic fidelity~(\ref{eq:fidelity}) exceeds 0.9999 as required by fault-tolerant quantum computing.
        The learning procedure output a successful policy.
		
	\subsection{Noise-resistant global optimization heuristics}
	
    DE is able to find feasible solutions in high-dimensional search space for a set of test problems~\cite{VT04} and for adaptive phase estimation~\cite{LCPS13}.
    However, when DE is employed for the problem of noisy phase estimation for $N$ up to 100, we observe that DE does not perform as well as PSO and, in fact, fails to deliver better than SQL scaling.
    To devise a noise-resistant global optimization algorithm for our scheme,
    we use the mean value $\bar{S}$ instead of $S$ (\ref{eq:Sharpness_est}) to determine the performance of a policy.
    This strategy is one of the many strategies proposed in the literature to create noise-resistant DE~\cite{DKC05,BPW04} and is found to work best for our problem.
    
    The principle behind the use of mean objective value is as follows: if noise is added to the fitness function,
    the process of averaging recovers the true objective value.
    The optimization using this value is, therefore, a close approximation to the noiseless optimization.
    The major drawback of this approach is that computing the objective function multiple times makes the procedure computationally expensive.
    Therefore, determining the smallest sample size of $\{S\}$ necessary to recover $S$ is crucial.
    To this end, we employ the heuristic applied to PSO in the previous work~\cite{HS11b}.
    The method updates $\bar{S}$ by computing one new sample of $S$ every iteration until a better offspring is generated.
    
    The sample size for computing $S$ is then determined by the probability for DE to generate an offspring that is better than the parent.
    This probability decreases as the candidates converge on the optimal value.
    As a result, the sample size grows automatically as the optimization progresses.
    The computational resources are allocated towards candidates that are close to optimal, which is a favorable strategy as the differences between objective values are dominated by noise in this case.
    Large sample size enables accurate estimations of the candidates' objective values.
    
    In the particular case of adaptive phase estimation, the phase noise is not additive in $S$ due to the exponential dependence.
    The mean value computed from $J$ samples, therefore, does not converge on the objective value of the noiseless case but an estimate of $S$ using sample size $JK$.
    This method, thereby, provides a better estimate of imprecision than for the sample size of $K$ for adaptive phase estimation including phase noise.
    
    \subsection{Improving scalability}
    In this subsection, we explain two techniques for achieving scalable learning algorithms.
One of the technique is to create an accept-reject criterion, allowing the algorithm to run for as long as it is necessary to generate a feasible policy.
    This technique is applied to the adaptive phase estimation, where successful policies from small values of $N$ are used to identify a region with a feasible policy for $N+1$.
    Another algorithm is devised for quantum gate design that alternates between optimizing in subspaces and overall space with self-adaptive DE.
    
	\subsubsection{Adaptive phase estimation at $N>90$}
	
In this subsubsection, we discuss accept-reject criteria and how this technique leads to DE delivering successful policies up to $N=100$.
	Although DE can generate successful policies for $N>45$, which is the limitation observed when PSO is used~\cite{LCPS13}, the variances also display stagnation when $N>90$.
	To generate policies from a search space that scales up to 100 dimensions, we implement a criterion to the noise-resistant DE to ensure that only successful policies are accepted.
	
	The stagnation is a manifestation of the algorithm not being able to converge to a successful solution in the time limit imposed.
	Previously the algorithm accepted a policy after a fixed number of iterations regardless of whether the population converges.
	However, as the dimension of the search space increases, so does the time for the population to converge.
	Eventually, the algorithm fails to deliver a policy that passes the test.
	We change the criterion for accepting a policy from a fixed number of iterations to only if $V_{\rm H}$ is within a distance corresponding to a confidence interval of 0.98 from the inverse power-law line.
	Thus, we guarantee that the policy from our algorithm always delivers a power-law scaling better than SQL.

The acceptable error $\delta_y$ for $N>93$ is calculated from the statistics of $V_{\rm H}$.
	The Holevo variance $V_{\rm H}$ are collected from $N=\{4,5,\dots,93\}$, in which we accept the policies after a fixed number of iterations.
	A linear equation is determined from $\{y_i\}=\{\log V_{\rm H}(N)\}$ and $\{x_i\}=\{\log N\}$, and is used to predict the next data point $y'$.
	The acceptable error from this predicted value is calculated using the previously stored data and the best value of $V_{\rm H}$ at iteration $G$ from a statistical formula, namely~\cite{YS09,BF10}
	{\small{
			\begin{equation}
				\delta_y=t^*_{n'-2}\sqrt{\frac{\sum_{i=1}^{n'}(y'_i-y_i)^2}{n'-2}\left(\frac{1}{n'}+\frac{(x'-\bar{x})^2}{\sum_{i=1}^{n'}(x_i-\bar{x})^2}\right)},
			\end{equation}}}where $n'$ is the number of data points, $x'=\log N$ for which the error is calculated, and $\bar{x}$ is the average of $\{x_i\}$.
			The value $t^*_{n'-2}$ is the quantile of the Student's $t$ distribution for $n'-2$ data points, which we approximate using a normal distribution.
			The policy is accepted if 
			\begin{equation}
			\left|\log V_{\rm H}(N)-y'\right|\leq\delta_y,
			\end{equation} or the optimization continues.
			
			The noise-resistant DE variant, including accept-reject criterion, works as follows\footnote[1]{Code available at~\url{http://panpalitta.github.io/phase_estimation/}}.
			\begin{algorithmic}
				\item \textbf{Step 1} Initialize the population of size $N_P$ randomly.
				\item \textbf{Step 2} Evaluate the objective function for each candidate \textit{twice}, and store the mean objective value and the sample size.
				\item \textbf{Step 3} Generate a donor $\bm{D}_i(G)$ for each of candidate $\bm{V}_i(G)$, where $G$ is the iterative time step, from three other candidates $\{\bm{V}_{i,1}(G),\bm{V}_{i,2}(G),\bm{V}_{i,3}(G)\}$ chosen randomly. For element $j$ of the donor $D_i(G)^{(j)}$,
						\begin{align}
							D_i(G)^{(j)}= \left\{
							\begin{array}{l l}
								V_{i,1}^{(j)}(G)+F\cdot(V_{i,2}^{(j)}(G)-V_{i,3}^{(j)}(G)),& \mathrm{if~}r\leq C_r,\\
								V_i(G)^{(j)},& \text{else},
							\end{array}\right.
						\end{align}
						where $F$ is the mutation rate, $C_r\in[0,1]$ is the crossover rate, and $r\in[0,1]$ is a random number.
						\item \textbf{Step 4} Evaluate the mean objective value for each of the new candidates from two samples.
						\item \textbf{Step 5} Compare and select the candidate for $G+1$ using the mean objective value,
						\begin{align}
							\bm{V}_i(G+1)= \left\{
							\begin{array}{l l}
								\bm{D}_i(G)& \text{if~} \bar{S}(\bm{D}_i(G))>\bar{S}(\bm{V}_i(G)),\\
								\bm{V}_i(G)& \text{else}.
							\end{array}
							\right.
						\end{align}
						\item \textbf{Step 6} Evaluate the objecting function once, and update the mean value and the sample size.
						\item \textbf{Step 7} Repeat steps 3 to 6 until the criterion to terminate the algorithm is met.
						\item \textbf{Step 8} Compute the objective value of the entire population 10 more times before selecting the candidate with the highest mean objective value as the solution.
					\end{algorithmic}
					
			The computational complexity of the algorithm is polynomial~\cite{LCPS13}, but it has a high degree, and therefore it is important to identify the performance critical parts of the implementation. We establish that over 90\% of the execution time is spent on generating random numbers one by one. 
			The random number generation is primarily used in estimating the Holevo variance as the computation involves simulations of the adaptive measurement procedure. 
			Generating random numbers as they are needed is not efficient on contemporary hardware. 
			The operations can be vectorized to use the single-instruction multiple-data architectures of the central and the graphics processing units (GPUs). 
			Abstracting the random number generation routines and introducing a buffer, we are able to vectorize the respective operations. 
			We study two approaches: one relies on the CPU, using the Intel Vector Statistical Library (VSL), the other on graphics processing units. Eventually, the VSL-based vectorized solution proves to be scalable.					
					
					\subsubsection{SuSSADE for quantum gate design}
					
                    In this subsubsection, we turn to another technique to improve DE's scalability and devise the subspace-selective self-adaptive differential evolution (SuSSADE).
                    SuSSADE combines two technique to improve convergence of the algorithm to a successful policy: the self-adaptive heuristic search and the reduction of the search space size.
                    The algorithm alternates between optimizing over the entire search space and one of the subspaces randomly selected for each iteration.
                                        
                   The efficacy of the algorithm for a particular landscape and dimension is determined by the search parameters ($F$, $C_r$), set to constant values in traditional DE. 
                    Instead of determining optimal values of the parameters through trial and error, which is infeasible when the problem has a large size such as in quantum gate design,
                    we implemented an algorithm to adapt the value of $F$ and $C_r$ between iterations.
                    This self-adaptive approach~\cite{BZM06} has also been used to enhance the performance of DE for the high-dimensional optimization problems and reads as follows.
                    
                    At iteration $G$, the algorithm determines the mutation rate $F$ and crossover rate $C_r$ for $G+1$ from                    
                    \begin{equation}
                        \label{Frate}
                        F_{G+1}= \begin{cases}
                            F_l+r_1\cdot F_u &\text{if $r_2<\kappa_1$}\\
                            F_{G} &\text{otherwise}\\
                        \end{cases}
                    \end{equation}
                    and
                    \begin{equation}
                        \label{Crate}
                        C_{r_{G+1}}= \begin{cases}
                            r_3 &\text{if $r_4<\kappa_2$}\\
                            C_{r_{G}} &\text{otherwise,}\\
                        \end{cases}
                    \end{equation}
                    where $r_j, j\in{1,2,3,4}$ are random numbers uniformly sampled from $\left(0,1\right]$.
                   The value of $F_l$ and $F_u$ are predetermined to be $0.1$ and $0.9$ respectively. The adaptive rate $\kappa_1$ and $\kappa_2$ are both set to $0.1$.
                    
                   In addition to adapting the search parameters, the algorithm also randomly decides whether the optimization is performed on the entire search space or a subspace. The switching rate $\mathcal{S}$ determines how often this switch occurs and is set by the user. DE is applied on the whole domain if a random number $r<\mathcal{S}$ for $r\in[0,1]$. Otherwise, the optimization is applied to a smaller domain of the problem.				
					Further details of the SuSSADE optimization algorithm\footnote[2]{Code available at~\url{https://github.com/ezahedin/DE_high-dimensional}} can be found in~\cite{ZGS15}.
					\section{Results}
					\label{sec:results}
					\subsection{AQEM}
					\begin{figure}
						\centering
						\includegraphics[scale=0.4]{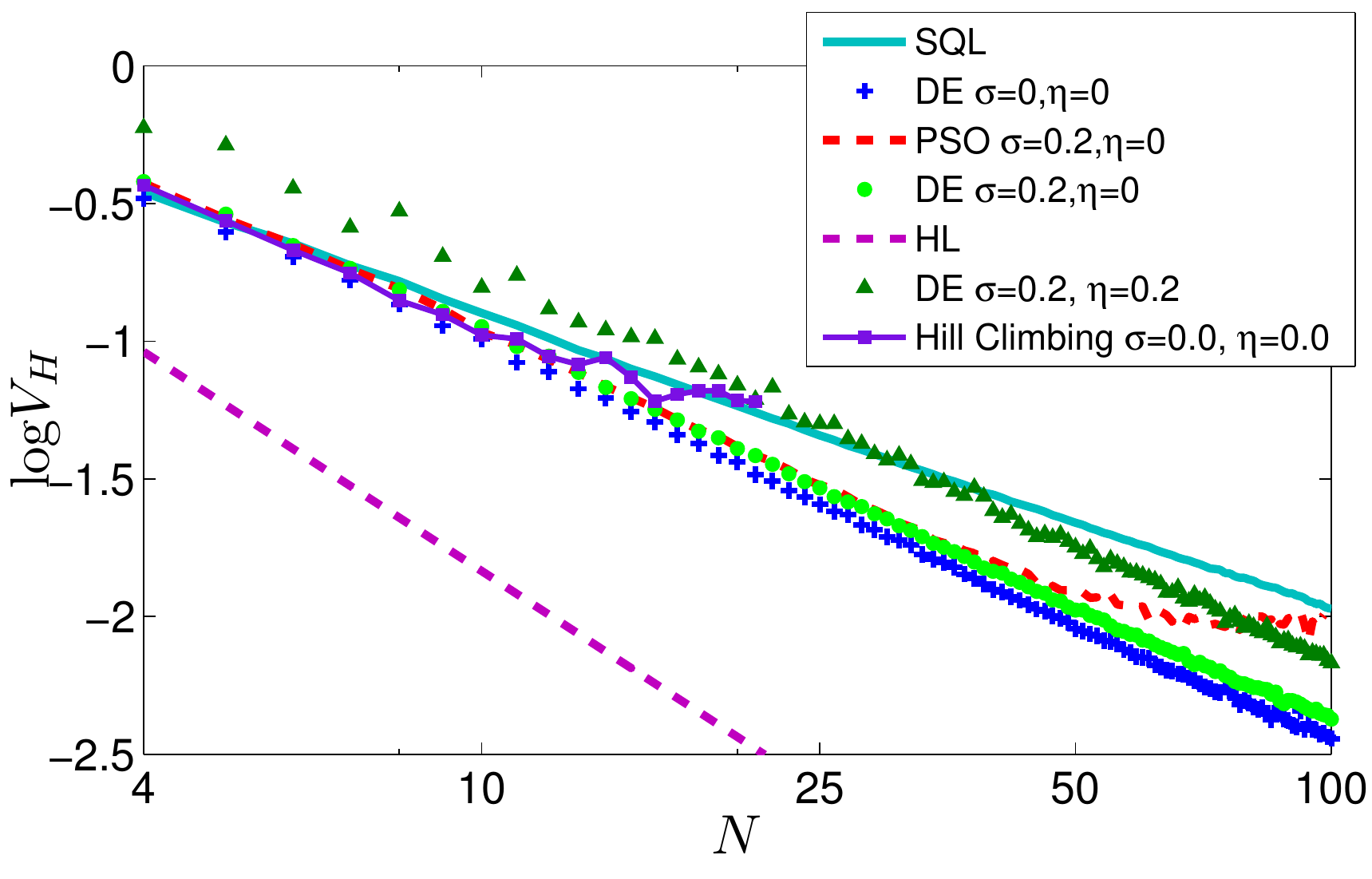}
						\caption{Logarithm of Holevo variance from adaptive interferometric-phase estimation. The interferometer includes small phase noise of width $\sigma$ and loss rate $\eta$. Three algorithms are used to generate the feedback policies: DE, PSO, and stochastic hill-climbing. This image is a rescaled version of Figure 1 in Ref.~\cite{PWS16}.}
						\label{fig:logHV}
					\end{figure}
					
                    By applying the method of creating noise-resistant to DE, we are able the obtain a policy that delivers the scaling of $V_H \propto N^{-1.421}$ when the width of the Gaussian distribution $\sigma=0.2$~rad, and the probability of losing a photon $\eta = 0.2$ are included. 
                    This result shows a scaling exceeding $N^{-1}$ expected from SQL, which is given for the ideal interferometer as a benchmark in Figure~\ref{fig:logHV}.
                    The SQL data is generated using a non-entangled $N$-particle state.
                    The HL shown is an extrapolation using the intercept from the SQL data and is included for the purpose to providing a possible benchmark for the scheme.
            
                    Although both the SQL and the HL are reported in the literature for the mean-squared error $\Delta\widetilde{\phi}$~\cite{GLM11}, we use the same benchmark for Holevo variance $V_{\rm H}$.
                    This follows from the approximation of $V_{\rm H}$ at low error $\left|\phi_k-\widetilde{\phi}_k\right|\ll 1$.
                    Under this condition, the sharpness in Eq.~(\ref{eq:Sharpness_est}) is approximated by a series expansion, and through this approximation, $V_{\rm H}$ is found to be the mean-squared error.
                    
                    The accept-reject criterion applied to $N>93$ enables the scheme to show the attain the power-law up to $N=100$ (Figure~\ref{fig:logHV}). 
                    The limitation at 100 photons is due to the computational time and the rounding error in the generation of the large multiparticle entangled state. The time required to find a policy under the accept-reject criterion from 94 to 100 photons is between 1.5 to 3 hours per data point.
                    
                    Policies that are found using stochastic hill climbing break down at 20 photons even for ideal phase estimation. The noise-resistant PSO shows the breakdown at 45 photons, consistent with the previous result~\cite{LCPS13}. We did not apply the accept-reject criterion to PSO as the computational time would have exceeded the time used by DE at the same number of $N$ and hence not considered worth an investment.
					
					\subsection{Quantum Gate Design}
\begin{figure}
					\centering
					    \includegraphics[scale=0.35]{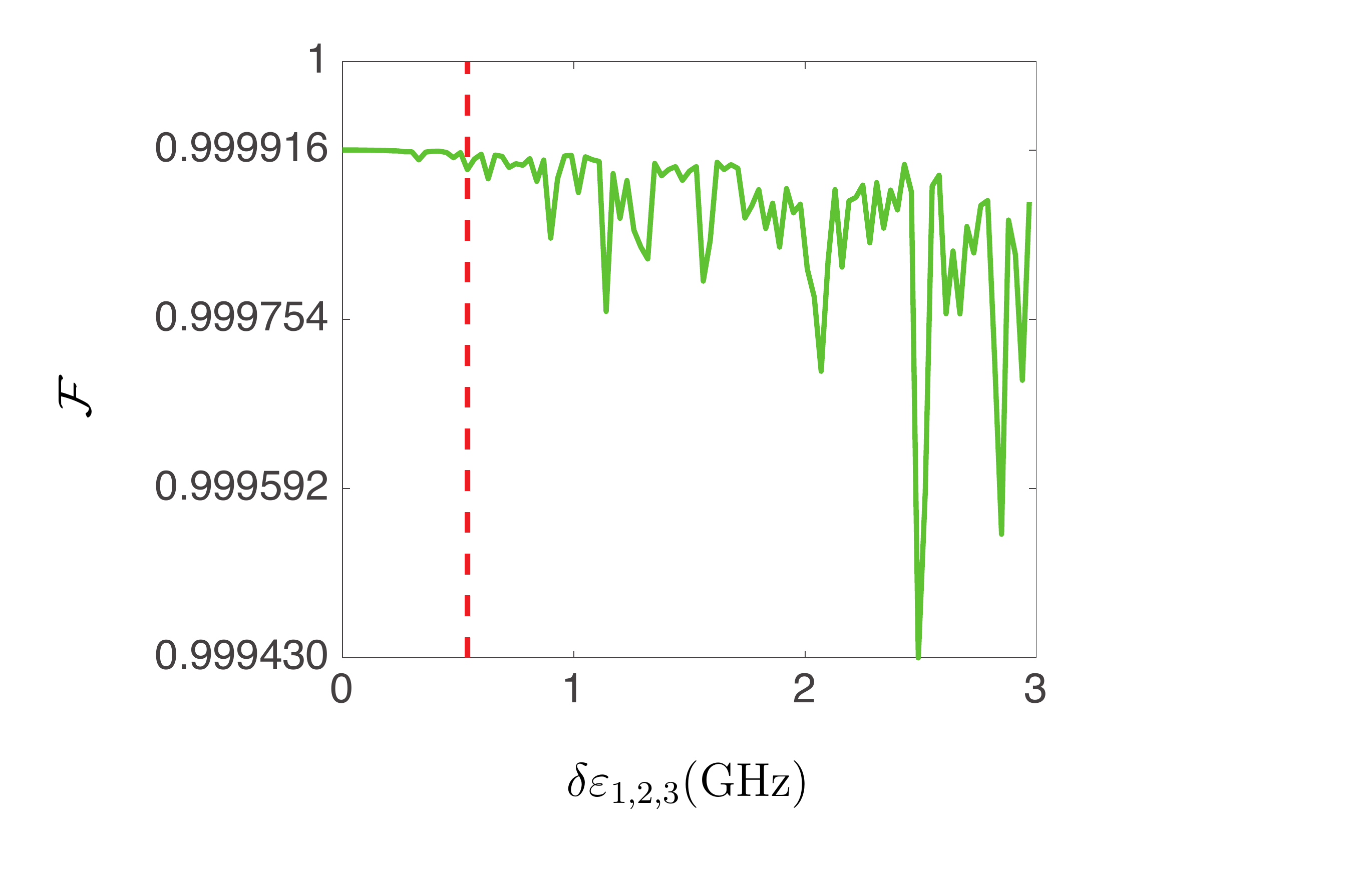}
					\caption{Intrinsic fidelity   $\mathcal{F}$ of the Toffoli gate as a function of $\delta\varepsilon_{1,2,3}$, corresponding to the noise level on each of the three transmons.
					The vertical red dotted line denotes the threshold, such that $\mathcal{F}>0.9999$ on the left of the line.}
					\label{fig:T_robustness}
					\end{figure}

                    Our machine learning approach to designing a three-qubit gate succeeds in generating policies for the design of a high-fidelity Toffoli gate. The resultant fidelity exceeds 0.9999, which exceeds the threshold fidelity for the fault-tolerant quantum computing. 
                    The gate operation time is found to be $26~\text{ns}$.
                    Therefore, the number of learning parameters add to the total of 81.
                    Although our machine learning technique has optimized the shape of the tranmons' frequencies over a piecewise-error-function, we have shown~\cite{ZGS15} that the algorithm does not rely on the shape of the pulse but on the number of learning parameters to generate successful policy for the gate design. 
                    Our quantum Toffoli gate operates as fast as a two-qubit entangling quantum gate under the same experimental conditions. 
                    The policies are also robust against the random uniform noise on the control pulses.
                    The threshold of the frequency $\delta\varepsilon$ of which the intrinsic fidelity remains above 0.9999 is well above the practical noise (up to 100kHz) of the control devices (Fig.~\ref{fig:T_robustness}).
					
					\section{Conclusion}
					\label{sec:conclusion}
                    In this work, we report on two examples of applying machine-learning algorithms to quantum control, namely the adaptive phase estimation and quantum gate design.
                    We employ reinforcement learning to adaptive phase estimation including noise and loss. We are able to attain enhanced precision better than SQL up to 100 photons using a noise-resistant variant of DE and accept-reject criterion.
                    The supervised-learning technique using SuSSADE enables us to perform single-shot high-fidelity three-qubit gates that are as fast as an entangling two-qubit gate under the same experimental constraints.
                    
                    The methods we employed do not require explicit knowledge of the system's dynamics, although the convexity of the objective functions, the dimension of the problems and the presence of noise have to be taken into account in order to generate a feasible policy.
                    We minimize the runtime of the algorithms by vectorizing the random number generation and employing GPUs and VSL.
                    This technique mostly affects the simulation of the quantum system as the simulations are the most time- and resource-consuming components of the current algorithms. 
                    
                    In principle, the simulation in the learning algorithms can be replaced by signals from experimental setup or simulations of other quantum control schemes.
                    This work can be used as the basis to develop learning algorithms for solving other quantum control problems, such as estimating more than one unknown parameters, which has an application in the characterization of quantum information processing devices, controlled quantum-state transfer in a spin chain~\cite{GB07}, and quantum error correction~\cite{ADL02}.
					
					\section{Acknowledgement}
					\label{sec:ackowledgement}
					This work is financially supported by NSERC and AITF. P.W. acknowledges financial support from the ERC (Consolidator Grant QITBOX), MINECO (Severo Ochoa grant SEV-2015-0522 and FOQUS), Generalitat de Catalunya (SGR 875), and Fundaci\'o Privada Cellex. B.C.S. also acknowledges support from the 1000 Talent Plan.
					The computational work is enabled by the support of WestGrid (www.westgrid.ca) and Calcul Qu\'ebec (www.calculquebec.ca) through Compute Canada Calcul Canada (www.computecanada.ca).
					
					\bibliographystyle{unsrt}
					\bibliography{noisyde}

\begin{thebibliography}{10}

\bibitem{Cav15}
C.~M. Caves.
\newblock Quantum information science: Emerging no more.
\newblock In P.~Kelley, G.~Agrawal, M.~Bass, J.~Hecht, and C.~Stroud, editors,
  {\em OSA Century of Optics}, pages 320--323. OSA, Washington, D. C., 2015.

\bibitem{Shor97}
P.~W. Shor.
\newblock Polynomial-time algorithms for prime factorization and discrete
  logarithms on a quantum computer.
\newblock {\em SIAM Journal on Computing}, 26(5):1484--1509, 1997.

\bibitem{Grov96}
L.~K. Grover.
\newblock A fast quantum mechanical algorithm for database search.
\newblock In G.~L. Miller, editor, {\em Proceedings of the Twenty-eighth Annual
  ACM Symposium on Theory of Computing}, STOC '96, pages 212--219, New York,
  May 1996. ACM, ACM.

\bibitem{BB84}
C.~H Bennett and G.~Brassard.
\newblock Quantum cryptography: {P}ublic key distribution and coin tossing.
\newblock In {\em IEEE International Conference on Computers, Systems and
  Signal Processing}, volume 175, pages 8--12, Piscataway, New Jersey, Dec
  1984. IEEE, IEEE.

\bibitem{SBC+09}
V.~Scarani, H.~Bechmann-Pasquinucci, Nicolas~J. Cerf,
  M.~Du\ifmmode~\check{s}\else \v{s}\fi{}ek, N.~L{\"u}tkenhaus, and M.~Peev.
\newblock The security of practical quantum key distribution.
\newblock {\em Reviews of Modern Physics}, 81:1301--1350, Sep 2009.

\bibitem{GLM11}
V.~Giovannetti, S.~Lloyd, and L.~Maccone.
\newblock Advances in quantum metrology.
\newblock {\em Nature Photonics}, 5(4):222--229, Mar 2011.

\bibitem{TA14}
G.~T\'{o}th and I.~Apellaniz.
\newblock Quantum metrology from a quantum information science perspective.
\newblock {\em Journal of Physics A: Mathematical and Theoretical},
  47(42):424006, Oct 2014.

\bibitem{NMR1}
C.~A. Ryan, M.~Laforest, J.~C. Boileau, and R.~Laflamme.
\newblock Experimental implementation of a discrete-time quantum random walk on
  an {NMR} quantum-information processor.
\newblock {\em Physical Review A}, 72(6):062317, 2005.

\bibitem{KRK+05}
N.~Khaneja, T.~Reiss, C.~Kehlet, T.~Schulte-Herbr\"{u}ggen, and S.~J. Glaser.
\newblock Optimal control of coupled spin dynamics: design of {NMR} pulse
  sequences by gradient ascent algorithms.
\newblock {\em Journal of Magnetic Resonance}, 172(2):296--305, 2005.

\bibitem{RSK05}
R.~Ruskov, K.~Schwab, and A.~N. Korotkov.
\newblock Squeezing of a nanomechanical resonator by quantum nondemolition
  measurement and feedback.
\newblock {\em Physical Review B}, 71(23):235407, 2005.

\bibitem{MVT98}
S.~Mancini, D.~Vitali, and P.~Tombesi.
\newblock Optomechanical cooling of a macroscopic oscillator by homodyne
  feedback.
\newblock {\em Physical Review Letters}, 80:688--691, Jan 1998.

\bibitem{HJH+03}
A.~Hopkins, K.~Jacobs, S.~Habib, and K.~Schwab.
\newblock Feedback cooling of a nanomechanical resonator.
\newblock {\em Physical Review B}, 68(23):235328, 2003.

\bibitem{ABB+98}
A.~Assion, T.~Baumert, M.~Bergt, T.~Brixner, B.~Kiefer, V.~Seyfried,
  M.~Strehle, and G.~Gerber.
\newblock Control of chemical reactions by feedback-optimized phase-shaped
  femtosecond laser pulses.
\newblock {\em Science}, 282(5390):919--922, 1998.

\bibitem{MS98}
D.~Meshulach and Y.~Silberberg.
\newblock Coherent quantum control of two-photon transitions by a femtosecond
  laser pulse.
\newblock {\em Nature}, 396(6708):239--242, 1998.

\bibitem{BS92}
P.~Brumer and M.~Shapiro.
\newblock Laser control of molecular processes.
\newblock {\em Annual Review of Physical Chemistry}, 43(1):257--282, 1992.

\bibitem{TR85}
D.~J. Tannor and S.~A. Rice.
\newblock Control of selectivity of chemical reaction via control of wave
  packet evolution.
\newblock {\em The Journal of Chemical Physics}, 83:5013, 1985.

\bibitem{DP10}
D.~Dong and I.~R. Petersen.
\newblock Quantum control theory and applications: A survey.
\newblock {\em IET Control Theory \& Applications}, 4(12):2651--2671, 2010.

\bibitem{SBW92}
R.~S. Sutton, A.~G. Barto, and R.~J. Williams.
\newblock Reinforcement learning is direct adaptive optimal control.
\newblock {\em IEEE Control Systems}, 12(2):19--22, Apr 1992.

\bibitem{KLM96}
L.~P. Kaelbling, M.~L. Littman, and A.~W. Moore.
\newblock Reinforcement learning: A survey.
\newblock {\em Journal of Artificial Intelligence Research}, 4:237--285, 1996.

\bibitem{KPK+04}
C.~P. Koch, J.~P. Palao, R.~Kosloff, and F.~Masnou-Seeuws.
\newblock Stabilization of ultracold molecules using optimal control theory.
\newblock {\em Physical Review A}, 70:013402, Jul 2004.

\bibitem{JRG+14}
G.~J{\"a}ger, D.~M. Reich, M.~H. Goerz, Christiane~P. Koch, and U.~Hohenester.
\newblock Optimal quantum control of {B}ose-{E}instein condensates in magnetic
  microtraps: {C}omparison of gradient-ascent-pulse-engineering and {K}rotov
  optimization schemes.
\newblock {\em Physical Review A}, 90:033628, Sep 2014.

\bibitem{RJ12}
B.~Rowland and J.~A. Jones.
\newblock Implementing quantum logic gates with gradient ascent pulse
  engineering: principles and practicalities.
\newblock {\em Philosophical Transactions of the Royal Society of London A:
  Mathematical, Physical and Engineering Sciences}, 370(1976):4636--4650, 2012.

\bibitem{AJSD+02}
M.~A. Armen, J.~K. Au, J.~K. Stockton, A.~C. Doherty, and H.~Mabuchi.
\newblock Adaptive homodyne measurement of optical phase.
\newblock {\em Physical Review Letters}, 89:133602, Sep 2002.

\bibitem{WBB+09}
H.~M. Wiseman, D.~W. Berry, S.~D. Bartlett, B.~L. Higgins, and G.~J. Pryde.
\newblock Adaptive measurements in the optical quantum information laboratory.
\newblock {\em IEEE Journal of Selected Topics in Quantum Electronics},
  15(6):1661--1672, Nov 2009.

\bibitem{Cap12}
P.~Cappellaro.
\newblock Spin-bath narrowing with adaptive parameter estimation.
\newblock {\em Physical Review A}, 85(3):030301, Mar 2012.

\bibitem{MV07}
M.~Mirrahimi and R.~{Van Handel}.
\newblock Stabilizing feedback controls for quantum systems.
\newblock {\em SIAM Journal on Control and Optimization}, 46(2):445--467, April
  2007.

\bibitem{VMS+12}
R.~Vijay, C.~Macklin, D.~H. Slichter, S.~J. Weber, K.~W. Murch, R.~Naik, A.~N.
  Korotkov, and I.~Siddiqi.
\newblock Stabilizing {R}abi oscillations in a superconducting qubit using
  quantum feedback.
\newblock {\em Nature}, 490(7418):77--80, Dec 2012.

\bibitem{WK98}
H.~M. Wiseman and R.~B. Killip.
\newblock Adaptive single-shot phase measurements: The full quantum theory.
\newblock {\em Physical Review A}, 57(3):2169--2185, Mar 1998.

\bibitem{BW00}
D.~W. Berry and H.~M. Wiseman.
\newblock Optimal states and almost optimal adaptive measurements for quantum
  interferometry.
\newblock {\em Physical Review Letters}, 85(24):5098--5101, Dec 2000.

\bibitem{WMW02}
H.~M. Wiseman, S.~Mancini, and J.~Wang.
\newblock Bayesian feedback versus markovian feedback in a two-level atom.
\newblock {\em Physical Review A}, 66:013807, Jul 2002.

\bibitem{RPH2015}
S.~Roy, I.~R. Petersen, and E.~H. Huntington.
\newblock Robust adaptive quantum phase estimation.
\newblock {\em New Journal of Physics}, 17(6):063020, Jun 2015.

\bibitem{TGB15}
M.~Tiersch, E.~J. Ganahl, and H.~J. Briegel.
\newblock Adaptive quantum computation in changing environments using
  projective simulation.
\newblock {\em Scientific Reports}, 5:12874, Aug 2015.

\bibitem{BPB15}
L.~Banchi, N.~Pancotti, and S.~Bose.
\newblock Quantum gate learning in engineered qubit networks: {T}offoli gate
  with always-on interactions.
\newblock {\em arXiv:1509.04298}, 2015.

\bibitem{AN16}
M.~August and X.~Ni.
\newblock Using recurrent neural networks to optimize dynamical decoupling for
  quantum memory.
\newblock {\em arXiv:1604.00279}, Apr 2016.

\bibitem{WEH+15}
P.~B. Wigley, P.~J. Everitt, A.~van~den Hengel, J.~W. Bastian, M.~A.
  Sooriyabandara, G.~D. McDonald, K.~S. Hardman, C.~D. Quinlivan, P.~Manju,
  C.~C.~N. Kuhn, I.~R. Petersen, A.~Luiten, J.~J. Hope, N.~P. Robins, and M.~R.
  Hush.
\newblock Fast machine-learning online optimization of ultra-cold-atom
  experiments.
\newblock {\em Scientific Reports}, 6:25890, May 2016.

\bibitem{HS10}
A.~Hentschel and B.~C. Sanders.
\newblock Machine learning for precise quantum measurement.
\newblock {\em Physical Review Letters}, 104(6):063603, Feb 2010.

\bibitem{LCPS13}
N.~B. Lovett, C.~Crosnier, M.~Perarnau-Llobet, and B.~C. Sanders.
\newblock Differential evolution for many-particle adaptive quantum metrology.
\newblock {\em Physical Review Letters}, 110(22):220501, May 2013.

\bibitem{Bis06}
C.~M. Bishop.
\newblock {\em {Pattern Recognition and Machine Learning}}.
\newblock Springer, Singapore, 2006.

\bibitem{MGM+15}
E.~Magesan, J.~M. Gambetta, A.~D. C\'orcoles, and J.~M. Chow.
\newblock Machine learning for discriminating quantum measurement trajectories
  and improving readout.
\newblock {\em Physical Review Letters}, 114:200501, May 2015.

\bibitem{MW10}
M.~Gu\c{t}\u{a} and W.~Kot\l{l}owski.
\newblock Quantum learning: asymptotically optimal classification of qubit
  states.
\newblock {\em New Journal of Physics}, 12(12):123032, 2010.

\bibitem{ZSS14}
E.~Zahedinejad, S.~Schirmer, and B.~C. Sanders.
\newblock Evolutionary algorithms for hard quantum control.
\newblock {\em Physical Review A}, 90:032310, Sep 2014.

\bibitem{SD94}
N.~Srinivas and K.~Deb.
\newblock Muiltiobjective optimization using nondominated sorting in genetic
  algorithms.
\newblock {\em Evolutionary Computation}, 2(3):221--248, 1994.

\bibitem{BYW+97}
C.~J. Bardeen, V.~V. Yakovlev, K.~R. Wilson, S.~D. Carpenter, P.~M. Weber, and
  W.~S. Warren.
\newblock Feedback quantum control of molecular electronic population transfer.
\newblock {\em Chemical Physics Letters}, 280(1-2):151--158, 1997.

\bibitem{SP97}
R.~Storn and K.~Price.
\newblock Differential evolution: A simple and efficient heuristic for global
  optimization over continuous spaces.
\newblock {\em Journal of Global Optimization}, 11(4):341--359, Dec 1997.

\bibitem{VT04}
J.~Vesterstrom and R.~Thomsen.
\newblock A comparative study of differential evolution, particle swarm
  optimization, and evolutionary algorithms on numerical benchmark problems.
\newblock In {\em Evolutionary Computation, 2004. CEC2004. Congress on},
  volume~2, pages 1980 -- 1987, Piscataway, New Jersey, Jun 2004. IEEE.

\bibitem{PSL05}
K.~V. Price, R.~M. Storn, and J.~A. Lampinen.
\newblock Benchmarking differential evolution.
\newblock In {\em Differential Evolution}, Natural Computing Series, pages
  135--187. Springer, Berlin, 2005.

\bibitem{Wis95}
H.~M. Wiseman.
\newblock Adaptive phase measurements of optical modes: Going beyond the
  marginal $q$ distribution.
\newblock {\em Physical Review Letters}, 75(25):4587--4590, Dec 1995.

\bibitem{WK97}
H.~M. Wiseman and R.B. Killip.
\newblock Adaptive single-shot phase measurements: A semiclassical approach.
\newblock {\em Physical Review A}, 56(1):944--957, Jul 1997.

\bibitem{OIO+12}
R.~Okamoto, M.~Iefuji, S.~Oyama, K.~Yamagata, H.~Imai, A.~Fujiwara, and
  S.~Takeuchi.
\newblock Experimental demonstration of adaptive quantum state estimation.
\newblock {\em Physical Review Letters}, 109:130404, Sep 2012.

\bibitem{ZGS15}
E.~Zahedinejad, Joydip G., and B.~C. Sanders.
\newblock High-fidelity single-shot {T}offoli gate via quantum control.
\newblock {\em Physical Review Letters}, 114:200502, May 2015.

\bibitem{ZGS16}
E.~Zahedinejad, J.~Ghosh, and B.~C. Sanders.
\newblock Designing high-fidelity single-shot three-qubit gates: A machine
  learning approach.
\newblock {\em arXiv:1511.08862v2}, Oct 2016.

\bibitem{DHJ+00}
A.~C. Doherty, S.~Habib, K.~Jacobs, H.~Mabuchi, and S.~M. Tan.
\newblock Quantum feedback control and classical control theory.
\newblock {\em Physical Review A}, 62:012105, Jun 2000.

\bibitem{AT12}
C.~Altafini and F.~Ticozzi.
\newblock Modeling and control of quantum systems: An introduction.
\newblock {\em IEEE Transactions on Automatic Control}, 57(8):1898--1917, Aug
  2012.

\bibitem{Wat11}
J.~Watrous.
\newblock Lecture notes on theory of quantum information.
\newblock University of Waterloo, Sep 2011.

\bibitem{Hol12}
A.~S. Holevo.
\newblock Quantum evolutions and channels.
\newblock In {\em Quantum Systems, Channels, Information: A Mathematical
  Introduction}, volume~16 of {\em De Gruyter Studies in Mathematical Physics},
  monograph Quantum evolutions and channels, pages 103--131. Walter de Gruyter,
  Berlin, dec 2012.

\bibitem{DFM-Z08}
J.~Dolbeault, P.~Felmer, and J.~Mayorga-Zambrano.
\newblock Compactness properties for trace-class operators and applications to
  quantum mechanics.
\newblock {\em Monatshefte f{\"u}r Mathematik}, 155(1):43--66, Sep 2008.

\bibitem{Bra03}
H.~E. Brandt.
\newblock Quantum measurement with a positive operator-valued measure.
\newblock {\em Journal of Optics B: Quantum and Semiclassical Optics},
  5(3):S266, Jun 2003.

\bibitem{Zur07}
W.~H. Zurek.
\newblock {\em Decoherence and the Transition from Quantum to Classical ---
  Revisited}, volume~48 of {\em Progress in Mathematical Physics}, pages 1--31.
\newblock Birkh{\"a}user Basel, Basel, 2007.

\bibitem{DPS03}
G.~M. D'Ariano, M.~G.~A. Paris, and M.~F. Sacchi.
\newblock Quantum tomography.
\newblock {\em Advances in Imaging and Electron Physics}, 128:206--309, 2003.

\bibitem{PWS16}
P.~Palittapongarnpim, P.~Wittek, and B.~C. Sanders.
\newblock Controlling adaptive quantum-phase estimation with scalable
  reinforcement learning.
\newblock In M.~Verleysen, editor, {\em 24th European Symposium on Artificial
  Neural Networks, Computational Intelligence and Machine Learning}, pages
  327--332, Belgium, April 2016. Ciaco.

\bibitem{ZPK12}
M.~Zwierz, C.~A. {P\'erez-Delgado}, and P.~Kok.
\newblock Ultimate limits to quantum metrology and the meaning of the
  {H}eisenberg limit.
\newblock {\em Physical Review A}, 85(4):042112, Apr 2012.

\bibitem{BWB01}
D.~W. Berry, H.~M. Wiseman, and J.~K. Breslin.
\newblock Optimal input states and feedback for interferometric phase
  estimation.
\newblock {\em Physical Review A}, 63(5):053804, May 2001.

\bibitem{Cav81}
C.~M. Caves.
\newblock Quantum-mechanical noise in an interferometer.
\newblock {\em Physical Review D}, 23(8):1693--1708, Apr 1981.

\bibitem{AAA+16}
B.~P. Abbott et~al.
\newblock Observation of gravitational waves from a binary black hole merger.
\newblock {\em Physical Review Letters}, 116(6):061102, Feb 2016.

\bibitem{BS13}
J.~Borregaard and A.~S. S\o{}rensen.
\newblock Near-{H}eisenberg-limited atomic clocks in the presence of
  decoherence.
\newblock {\em Physical Review Letters}, 111(9):090801, Aug 2013.

\bibitem{Hil02}
J.~Hilgevoord.
\newblock The standard deviation is not an adequate measure of quantum
  uncertainty.
\newblock {\em American Journal of Physics}, 70(10):983--983, Oct 2002.

\bibitem{FA01}
G.~W. Forbes and M.~A. Alonso.
\newblock Measures of spread for periodic distributions and the associated
  uncertainty relations.
\newblock {\em American Journal of Physics}, 69(3):340--347, Mar 2001.

\bibitem{Shor96}
P.~W. Shor.
\newblock Fault-tolerant quantum computation.
\newblock In {\em Foundations of Computer Science, 1996. Proceedings., 37th
  Annual Symposium on}, pages 56--65, Piscataway, New Jersey, 1996. IEEE, IEEE.

\bibitem{SWS16}
Y.~R. Sanders, J.~J. Wallman, and B.~C. Sanders.
\newblock Bounding quantum gate error rate based on reported average fidelity.
\newblock {\em New Journal of Physics}, 18(1):012002, 2016.

\bibitem{BBC+95}
A.~Barenco, C.~H. Bennett, R.~Cleve, D.~P. DiVincenzo, N.~Margolus, P.~Shor,
  T.~Sleator, J.~A. Smolin, and H.~Weinfurter.
\newblock Elementary gates for quantum computation.
\newblock {\em Physical Review A}, 52(5):3457--3467, Nov 1995.

\bibitem{MVBS04}
M.~M{\"o}tt{\"o}nen, J.~J. Vartiainen, V.~Bergholm, and M.~M. Salomaa.
\newblock Quantum circuits for general multiqubit gates.
\newblock {\em Physical Review Letters}, 93(13):130502, Sep 2004.

\bibitem{FSB+12}
A.~Fedorov, L.~Steffen, M.~Baur, M.~P. da~Silva, and Andreas Wallraff.
\newblock Implementation of a {T}offoli gate with superconducting circuits.
\newblock {\em Nature}, 481(7380):170--172, January 2012.

\bibitem{MKH+09}
T.~Monz, K.~Kim, W.~H\"ansel, M.~Riebe, A.~S. Villar, P.~Schindler, M.~Chwalla,
  M.~Hennrich, and R.~Blatt.
\newblock Realization of the quantum {T}offoli gate with trapped ions.
\newblock {\em Physical Review Letters}, 102:040501, Jan 2009.

\bibitem{RDN+12}
M.~D. Reed, L.~DiCarlo, S.~E. Nigg, L.~Sun, L.~Frunzio, S.~M. Girvin, and R.~J.
  Schoelkopf.
\newblock Realization of three-qubit quantum error correction with
  superconducting circuits.
\newblock {\em Nature}, 482(7385):382--385, February 2012.

\bibitem{LBA+09}
B.~P. Lanyon, M.~Barbieri, M.~P. Almeida, T.~Jennewein, T.~C. Ralph, K.~J.
  Resch, G.~J. Pryde, J.~L. {O'Brien}, A.~Gilchrist, and A.~G. White.
\newblock Simplifying quantum logic using higher-dimensional {H}ilbert spaces.
\newblock {\em Nature Physics}, 5(2):134--140, February 2009.

\bibitem{SHK+08}
J.~A. Schreier, A.~A. Houck, Jens Koch, D.~I. Schuster, B.~R. Johnson, J.~M.
  Chow, J.~M. Gambetta, J.~Majer, L.~Frunzio, M.~H. Devoret, S.~M. Girvin, and
  R.~J. Schoelkopf.
\newblock Suppressing charge noise decoherence in superconducting charge
  qubits.
\newblock {\em Physical Review B}, 77:180502, May 2008.

\bibitem{Leigh04}
J.R. Leigh.
\newblock {\em Control Concepts: A Non-Mathematical Introduction}, chapter~1.
\newblock Institution of Engineering and Technology, London, UK., 2 edition,
  2004.

\bibitem{Haid13}
M.~A. Haidekker.
\newblock Introduction to linear feedback controls.
\newblock In Mark~A. Haidekker, editor, {\em Linear Feedback Controls},
  chapter~1, pages 1 -- 13. Elsevier, Oxford, UK, Aug 2013.

\bibitem{AM08}
K.~J. Astr{\"o}m and R.~M. Murray.
\newblock {\em Introduction}, chapter~1, pages 1--26.
\newblock Princeton university press, Princeton, NJ, 2008.

\bibitem{Vap95}
V.~Vapnik.
\newblock {\em The Nature of Statistical Learning Theory}.
\newblock Springer, New York, 1995.

\bibitem{DGL96}
L.~Devroye, L.~Gy\"orfi, and G.~Lugosi.
\newblock {\em A Probabilistic Theory of Pattern Recognition}.
\newblock Springer, New York, 1996.

\bibitem{Long99}
P.~M. Long.
\newblock The complexity of learning according to two models of a drifting
  environment.
\newblock {\em Machine Learning}, 37(3):337--354, 1999.

\bibitem{SB98}
Richard~S. Sutton and Andrew~G. Barto.
\newblock Problem.
\newblock In {\em Reinforcement Learning: An Introduction}, Adaptive
  Computation and Machine Learning, chapter~1. MIT, Massachusetts, 1998.

\bibitem{Kak03}
S.~M. Kakade.
\newblock {\em On the Sample Complexity of Reinforcement Learning}.
\newblock PhD thesis, University College London, 2003.

\bibitem{Li09}
L.~Li.
\newblock {\em A Unifying Framework for Computational Reinforcement Learning
  Theory}.
\newblock PhD thesis, Rutgers University, 2009.

\bibitem{EBG11}
S.~Ertekin, L.~Bottou, and C.~L. Giles.
\newblock Nonconvex online support vector machines.
\newblock {\em IEEE Transactions on Pattern Analysis and Machine Intelligence},
  33(2):368--381, Feb 2011.

\bibitem{KE95}
J.~Kennedy and R.~Eberhart.
\newblock Particle swarm optimization.
\newblock In {\em Neural Networks, 1995. Proceedings., IEEE International
  Conference on}, volume~4, pages 1942--1948, Piscataway, New Jersey, Nov 1995.
  IEEE.

\bibitem{CDL+2014}
C.~Chen, D.~Dong, H.~X. Li, J.~Chu, and T.~J. Tarn.
\newblock Fidelity-based probabilistic q-learning for control of quantum
  systems.
\newblock {\em IEEE Transactions on Neural Networks and Learning Systems},
  25(5):920--933, May 2014.

\bibitem{MCD2015}
H.~Ma, C.~Chen, and D.~Dong.
\newblock Differential evolution with equally-mixed strategies for robust
  control of open quantum systems.
\newblock In {\em Systems, Man, and Cybernetics (SMC), 2015 IEEE International
  Conference on}, pages 2055--2060, Piscataway, NJ, Oct 2015. IEEE.

\bibitem{WO12}
M.~{van Otterlo} and M.~Wiering.
\newblock {\em Reinforcement Learning and {M}arkov Decision Processes},
  volume~12 of {\em Adaptation, Learning, and Optimization}, pages 3--42.
\newblock Springer, Berlin, 2012.

\bibitem{HS11b}
A.~Hentschel and B.~C. Sanders.
\newblock Efficient algorithm for optimizing adaptive quantum metrology
  processes.
\newblock {\em Physical Review Letters}, 107(23):233601, Nov 2011.

\bibitem{DKC05}
S.~Das, A.~Konar, and U.~K. Chakraborty.
\newblock Improved differential evolution algorithms for handling noisy
  optimization problems.
\newblock In {\em 2005 IEEE Congress on Evolutionary Computation}, volume~2,
  pages 1691--1698, Piscataway, New Jersey, Sep 2005. IEEE.

\bibitem{BPW04}
L.~Barone, A.~{Di Pietro}, and L.~While.
\newblock Applying evolutionary algorithms to problems with noisy,
  time-consuming fitness functions.
\newblock In G.~Greenwood, editor, {\em Evolutionary Computation, 2004.
  CEC2004. Congress on}, volume~2, pages 1254--1267, Oregon, Jun 2004. IEEE.

\bibitem{YS09}
X.~Yan and X.~G. Su.
\newblock {\em Simple Linear Regression}, chapter~2, pages 9--40.
\newblock SG: World Scientific, Singapore, 2009.

\bibitem{BF10}
N.~H. Bingham and J.~M. Fry.
\newblock {\em Linear Regression}, chapter~1, pages 1--32.
\newblock Springer Undergraduate Mathematics. Springer, London, 2010.

\bibitem{BZM06}
J.~Brest, V.~Zumer, and M.~S. Maucec.
\newblock Self-adaptive differential evolution algorithm in constrained
  real-parameter optimization.
\newblock In {\em 2006 IEEE International Conference on Evolutionary
  Computation}, pages 215--222, Piscataway, New Jersey, 2006. IEEE.

\bibitem{GB07}
J.~Gong and P.~Brumer.
\newblock Controlled quantum-state transfer in a spin chain.
\newblock {\em Physical Review A}, 75:032331, Mar 2007.

\bibitem{ADL02}
C.~Ahn, A.~C. Doherty, and A.~J. Landahl.
\newblock Continuous quantum error correction via quantum feedback control.
\newblock {\em Physical Review A}, 65:042301, Mar 2002.

\end{thebibliography}
					
				\end{document}